\documentclass[journal,onecolumn,11pt]{IEEEtran}

\usepackage{bm}
\usepackage{amssymb}
\usepackage{bm} 
\usepackage{stfloats}
\usepackage{amsmath} 
\usepackage{dsfont} 
\usepackage{amssymb}
\usepackage{amsmath, amsthm}
\newtheorem{defn}{Definition}

\usepackage{bm}

\usepackage{amsmath, amssymb, amsfonts, amsbsy, mathrsfs, bm}
\usepackage{cite, enumerate}
\usepackage[table]{xcolor}
\usepackage[normalem]{ulem}
\usepackage{dsfont}

\newtheorem{rem}{Remark}

\usepackage{multicol}
\usepackage{subfigure}
\usepackage{amsmath}
\usepackage{amssymb}
\usepackage{amsfonts}
\usepackage{graphicx}
\usepackage{url}
\usepackage{booktabs} 
\usepackage{amssymb}
\usepackage{stfloats}
\usepackage{amssymb}
\usepackage{bm} 
\usepackage{stfloats}
\usepackage{amsmath} 
\usepackage{dsfont} 
\usepackage{subfigure}
\usepackage[table]{xcolor}

\usepackage{graphicx}

\begin{document}
%
\title{An Improved Self-Organizing Diffusion Mobile Adaptive Network for Pursuing a Target}
%
%
%

\author{Amir Rsategarnia, Azam Khalili, and Md Kafiul Islam
\thanks{Amir Rsategarnia and Azam Khalili are with the Department of Electrical Engineering, University of Malayer, Malayer 65719-95863, Iran, emails: {a\_rastegar@ieee.org, a.khalili}@ieee.org}
\thanks{Md Kafiul Islam is with Department of Electrical and Computer Engineering, National University of Singapore, Singapore-117583, Email: kafiul\_islam@nus.edu.sg}}

%

\markboth{Journal of \LaTeX\ Class Files,~Vol.~6, No.~1, January~2007}%
{Shell \MakeLowercase{\textit{et al.}}: Bare Demo of IEEEtran.cls for Journals}
%

\maketitle
\thispagestyle{empty}

\begin{abstract}

\small{In this letter we focus on designing self-organizing diffusion mobile adaptive networks where the individual agents are allowed to move in pursuit of an objective (target). The well-known Adapt-then-Combine (ATC) algorithm is already available in the literature as a useful distributed diffusion-based adaptive learning network. However, in the ATC diffusion algorithm, fixed step sizes are used in the update equations for velocity vectors and location vectors. When the nodes are too far away from the target, such strategies may require large number of iterations to reach the target. To address this issue, in this paper we suggest two modifications on the ATC mobile adaptive network to improve its performance. The proposed modifications include (i) distance-based variable step size adjustment at diffusion algorithms to update velocity vectors and location vectors (ii) to use a selective cooperation, by choosing the best nodes at each iteration to reduce the number of communications. The performance of the proposed algorithm is evaluated by simulation tests where the obtained results show the superior performance of the proposed algorithm in comparison with the available ATC mobile adaptive network.} 
\end{abstract}


\section{Introduction}
Wireless sensor networks appear in many practical applications such as distributed sensing, intrusion detection and target localization \cite{est01,leu08,yadav12, chao13}. In most of the aforementioned applications, nodes of a network collect data from environment and then process them colaborately to estimate a desired parameter. Different strategies have been introduced in the literature to solve the distributed estimation problems including consensus strategies and adaptive networks \cite{sayed14a,sayed14b}. It has been shown in \cite{tu12} that adaptive networks are more stable than consensus networks and they provide better steady-state error performance. So, in this paper we focus on adaptive network based solutions. We adopt the term adaptive networks from \cite{sayed06} to refer to a collection of nodes that interact with each other and function as a single adaptive entity that is able to track statistical variations of data in real-time. Two major classes of adaptive networks are incremental strategy \cite{lopes07a,tak08,pan12,azar15,khalili15a,bazzi15} and diffusion strategy \cite{lopes07b,cat10,chen12,sayed13,zhao15}. In comparison, incremental algorithms require less communication among nodes of the networks while diffusion algorithms are scalable and more robust to link and node failure \cite{kha12,rast14,zhao12}. In general, diffusion based algorithms consist of two steps including the adaptation step, where the node updates the weight estimate using local measurement data, and the combination step where the information from the neighbors are aggregated. Based on the order of these two steps, diffusion algorithms can be categorized into two classes known as the Combine-then-Adapt and Adapt-then-Combine (ATC). It is observed that the ATC version of diffusion LMS outperforms the CTA algorithm \cite{cat10}. 

The initial diffusion adaptive networks in \cite{lopes07b,cat10,chen12,sayed13,zhao15} did not incorporate the node mobility. In \cite{tu10a,tu10b,tu11,zar15}, another dimension of complexity which is node mobility has been added to the diffusion networks. The resultant mobile adaptive networks perform two diffusion-based estimation tasks: one for estimating the location of a target and the other one for tracking the center of mass of the network. Incorporating the node mobility enables the resulting diffusion networks to use them in new applications such as modeling the various forms of sophisticated behavior exhibited by biological networks \cite{cat11c,li12} and source localization \cite{li11,kha13}.

The current algorithms for mobile adaptive networks do not consider the \emph{distance to the target} in their adaptation mechanisms. In other words in the current algorithms every node in the network adjusts (updates) its velocity vector and location vector regardless of its distance to the target. When nodes are too far away from the target, such strategies may require large number of iterations to reach the target, which in turn, requires large amount of communications and computations. Thus, it is highly desirable that in a mobile adaptive network, the nodes incorporate the distance to the target information, e.g. using bigger step-sizes when they are too far from the target. To endow the ATC diffusion network with such ability, we firstly define a practical metric to describe the far field region as a region that is too far from the target\footnote{Note that the near filed is also defined as a region that includes the target.}. Then, according to the position of a node (inside or outside of the far field) different  step size adjust mechanisms are applied.  In general, as long as a node is inside the far field region, the  step size parameter in update velocity vectors and location vectors are increased, whereas for a node outside of the far field the step size is iteratively reduced as it moves toward the target.

Moreover, as we will further discuss in Section 2, in the combination step of diffusion LMS algorithm, each node needs to gather the intermediate estimates from all of its neighbours. Thus, this step may requires large amount of communications at each iteration for dense networks. In some applications, however, networks cannot afford large communication overhead. To further reduce the communication load, a selective cooperation is used where every nodes selects only a subset of its neioubours to share the information.  The performance of the proposed algorithm is evaluated by simulation tests where the obtained results show the superior performance of the proposed algorithm in comparison with the available ATC mobile adaptive network.

\textbf{Notation}: Throughout the paper we use boldface letters for matrices and vectors and small letters for scalars. The notation $\|\mathbf{x}\|^2=\mathbf{x}^* \mathbf{x}$ stands for the Euclidean norm of $\bf{x}$. $E$ denotes the expectation operator. The set of neighbors of
node $k$, including itself, is called the neighborhood of and is denoted by $\mathcal{N}_k$. The exact meaning of this notation will be clear from the context. 

\begin{figure}[t]
\centering 
\includegraphics [width=7.5cm]{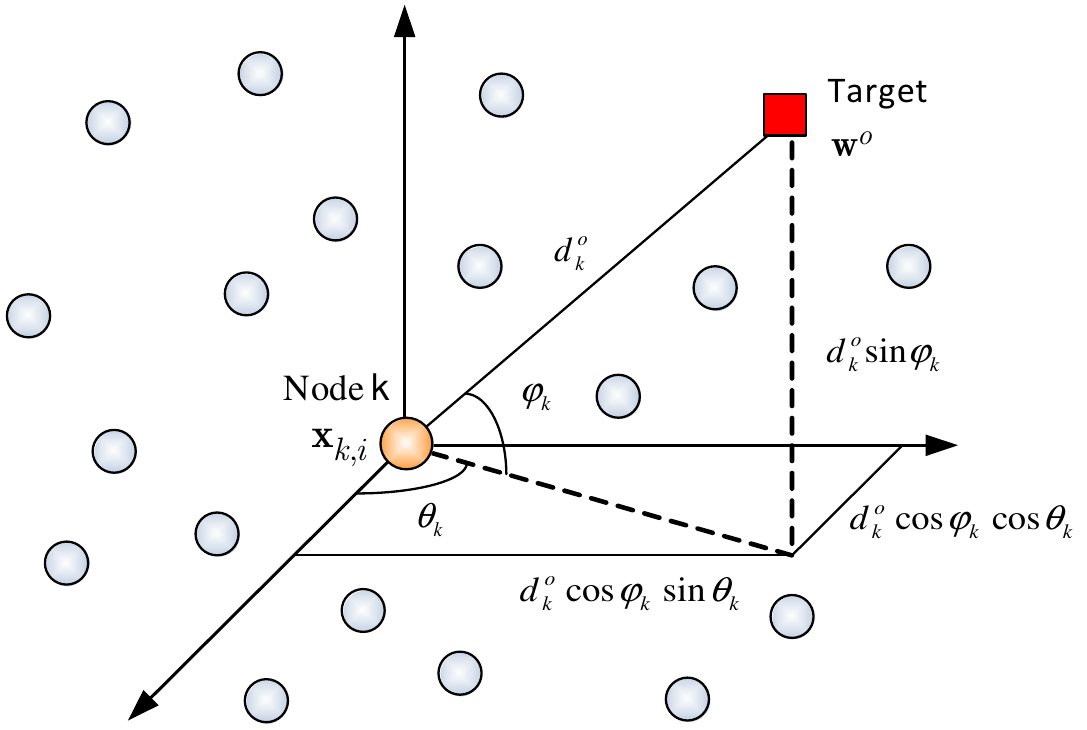} 
\centering\caption{Distance and direction of the target $\mathbf{w}^o$ from node $k$ at location $\mathbf{x}_{k,i}$.}
\label{fig-1}
\end{figure}

\section{Mobile adaptive networks}
Let us consider a network with $N$  mobile nodes that are randomly distributed over a space. Let $\mathbf{x}_{k,i}$ be the location of node $k$ at time $i$ (relative to some global coordinate system) and $\mathbf{w}^o \in \mathbb{R}^3$  denote the location of the target.  Each node $k$ finds its neighbors within a
range $R$ radius in each time $i$, i.e. 
\begin{equation} \label{ndef}
	j \in \mathcal{N}_k\ \ \mathrm{if}\ \ \|\mathbf{x}_{j,i}-\mathbf{x}_{k,i}\|\leq R
\end{equation}
The objectives of each node in the network are (1) to estimate the position of the desired target and move towards it in coherence and synchrony with
the other nodes, and (2), avoid possible collisions by keeping a certain distance, say $r$ from neighbors during the motion to the target location.
The distance between the target $\mathbf{w}^o$  and a node $k$ at any time $i$  is given by (See Fig. \ref{fig-1})
\begin{equation}
d_{k,i}^o = {{\mathbf{u}}_{k,i}}({{\mathbf{w}}^o} - {{\mathbf{x}}_{k,i}})
\end{equation}
where $\mathbf{u}_{k,i}$  denotes the direction of the target including the azimuth angle $\theta _{k,i}$, and the elevation angle $\varphi _{k,i}$ which is given as
\begin{equation}
{\mathbf{u}}_{k,i} = [ {\cos ({\theta _{k,i}})\cos ({\varphi _{k,i}})} \ \  {\sin ({\theta _{k,i}})\cos ({\varphi _{k,i}})} \ \  {\sin ({\varphi _{k,i}})} ]
\end{equation}
We assume that each node observes a noisy measurement of the distance to the target, via a linear model  as follows
\begin{equation}
d_{k,i}^n = {{\bf{u}}_{k,i}}({{\bf{w}}^o} - {{\bf{x}}_{k,i}}) + {n_{k,i}}
\end{equation}
where $n_{k,i}$ denotes the noise term which is assumed to be zero-mean Gaussian noise. Intuitively, the noise variance,
$\sigma_{n,k}^2(i)$ can follow a relation such as
\begin{equation} \label{noap}
\sigma_{n,k}^2(i) \approx \kappa \|\mathbf{x}_{k,i}-\mathbf{w}^o\|^2
\end{equation}
We select $\kappa=0.01$ as \cite{tu10a}. Note that \eqref{noap} is reasonable since we usually assume the signal power to decrease in
proportional to the square of the propagation distance. We can rewrite the above equation as \cite{tu10a,tu11}
\begin{equation}
{d_{k,i}} = d_{k,i}^n + {{\bf{u}}_{k,i}}{{\bf{x}}_{k,i}} = {{\bf{u}}_{k,i}}{{\bf{w}}^o} + {n_{k,i}}
\end{equation}
At every time instant $i$, every node $k$ has access to local data $\{d_{k,i},\mathbf{u}_{k,i}\}$ and the local data from its neighbors. Using these data every node estimates the position of the target at $\mathbf{w}^o$ which can be achieved by solving the following optimization problem
\begin{equation}  \label{cfw}
\arg \min_{\bf{w}}\big( \sum\limits_{k = 1}^N {E\left[ {{{\left| {{d_{k,i}} - {{\mathbf{u}}_{k,i}}{\mathbf{w}}} \right|}^2}} \right]}\big)
\end{equation}
The ATC diffusion algorithm has been developed for solving \eqref{cfw} in a distributed manner. The ATC algorithm consists of two steps: adaptation and combination. In adaptation step, node $k$ uses its own data to update the weight estimate $\mathbf{w}_{k,i-1}$ to intermediate
value $\mathbf{m}_{k,i}$. In the combination step each node gathers the intermediate estimates $\{\mathbf{m}_{\ell,i}\}, \ell \in \mathcal{N}_k$ combines them to obtain the updated weight estimate $\mathbf{w}_{k,i}$. The algorithm is described as follows:
\begin{align} \label{eqwup}
{{{\bf{m}}}_{k,i}} &= {{\bf{w}}_{k,i - 1}} + {\mu _{k}}\sum\limits_{\ell  \in {{\cal N}_k}} c_{\ell ,k}^w{\bf{u}}_{\ell ,i}^T\left( {{d_{\ell ,i}} - {{\bf{u}}_{\ell ,i}}{{\bf{w}}_{k,i - 1}}} \right)  \\ 
 {{\bf{w}}_{k,i}} &= \sum\limits_{\ell  \in {{\cal N}_k}} a_{\ell ,k}^w{{\bf{m}}_{\ell ,i}} 
\end{align}
where $\mu_{k}$ is the learning step size. The two sets of non-negative real coefficients $\{c_{\ell ,k}^w\}$ and $\{a_{\ell ,k}^w\}$ satisfy
\begin{align} \label{conds}
\sum_{\ell=1}^{N} c_{\ell ,k}^w&=\sum_{\ell=1}^{N} a_{\ell ,k}^w=1,\ \  \ \ c_{\ell ,k}^w=a_{\ell ,k}^w=0,\ \ \ell \notin \mathcal{N}_k
\end{align}

In a network of mobile nodes, every node $k$ can update its location as 
\begin{equation}
{\mathbf{x}_{k,i }} = {\mathbf{x}_{k,i-1}} + {\mathbf{v}_{k,i}} \Delta t 
\end{equation}
where $\Delta t$ is the time step and ${\mathbf{v}_{k,i + 1}}$ denoted the velocity of the node $k$. Every node adjusts its velocity vector according to the following expression \cite{tu10a}
\begin{align} \label{vfor}
{\mathbf{v}_{k,i}} &= \xi_1 \frac{{\mathbf{w}_{k,i-1}} - {\mathbf{x}_{k,i-1}}}{{\left\| {{\mathbf{w}_{k,i-1}} - {\mathbf{x}_{k,i-1}}} \right\|}} + \xi_2 \mathbf{v}_{k,i-1}^g   \nonumber   \\
& \hspace{1cm}+ \xi_3 \sum\limits_{l \in {\mathcal{N}_k}\backslash \{k\} }(\|\mathbf{x}_{l,i}-\mathbf{x}_{k,i}\|-r) \frac{\mathbf{x}_{l,i}-\mathbf{x}_{k,i}}{\|\mathbf{x}_{l,i}-\mathbf{x}_{k,i}\|}
\end{align}
where $\xi_1$, $\xi_2$  and  $\xi_3$ are non-negative weighting factors, and $\mathbf{v}_{k,i}^g$ is the local estimate for the global velocity of the center of gravity of the network which is designed to allow for coherent motion. To use \eqref{vfor} each node needs to estimate $\mathbf{v}_{k,i}^g$ in a distributed way. Since the velocities of nodes are changing in time, we need to keep track of ${\mathbf{v}}_{k,i}^g$ over time. So we introduce the global cost function as follows
\begin{equation}
\arg \min_{\mathbf{v}^g}\big( \sum\limits_{k = 1}^N {E\left[ {{{\left| {{{\mathbf{v}}_{k,i}} - {{\mathbf{v}}^g}} \right|}^2}} \right]}\big)
\end{equation}
Similar to \eqref{eqwup} we can arrive at the following diffusion algorithm for estimating ${\mathbf{v}}_{k,i}^g$
\begin{align}
{\mathbf{s}}_{k,i} &= {\mathbf{v}}_{k,i - 1}^g + {\delta_k}\sum\limits_{\ell  \in {{\cal N}_k}} {c_{\ell ,k}^v\left( {{{\mathbf{v}}_{\ell ,i}} - {\mathbf{v}}_{k,i - 1}^g} \right)} \\
{\mathbf{v}}_{k,i}^g &= \sum\limits_{\ell  \in {{\cal N}_k}} {a_{\ell ,k}^v{{{\mathbf{s}}}_{\ell ,i}}} 
\end{align}
where $\delta_k$ is a positive step size and $\{c_{\ell ,k}^v\}$ and $\{a_{\ell ,k}^v\}$  are two sets of non-negative real coefficients satisfying the same properties as \eqref{conds}.

\section{Proposed Algorithm}
\subsection{Motivation for current work}
In the existing algorithms for adaptive mobile networks, there is no specific strategy for going faster toward the target. In other words, in the existing algorithms although the algorithm is designed in such a way that the set of nodes can move towards the goal harmoniously, but they do not consider the \emph{distance to the target} in their velocity and location update equations. To address this issue, instead of using fixed learning parameter $\mu_k$ in \eqref{eqwup}, we use variable step-size parameter which has the following conditions
\begin{itemize}
	\item if  node $k$ at iteration $i$ is too far from the target, $\mu_{k,i}$ should be increased. Note that to prevent algorithm divergence and movement control of the set of nodes, we have to select the upper limit for step-size parameters.
	\item if distance of node $k$ at iteration $i$ becomes less than a predefined value, $\mu_{k,i}$ should be iteratively decreased as nodes approach the target. In this case we consider a lower bound for $\mu_{k,i}$ to avoid slow convergence rate. 
\end{itemize}

\subsection{Algorithm Development}
To begin with, firstly we define the far field region as follows.
\begin{defn}
By far field, we mean a region $\mathcal{D} \in \mathbb{R}^3$ that the nodes inside it are far from the target, or 
\begin{equation} \label{def1}
\|\mathbf{w}^o-\mathbf{x}_{k,i}\|^2 \gg s,
\end{equation}
where $s>0$.
\end{defn}
To use \eqref{def1}, every node needs to have the target position $\mathbf{w}^o$ which is not available. To have a practical metric, we use $\mathbf{w}_{k,i}$ as an estimate of $\mathbf{w}^o$ at iteration $i$ and rewrite  \eqref{def1} as
\begin{equation} \label{def2}
\|\mathbf{w}_{k,i}-\mathbf{x}_{k,i}\|^2 \gg s,
\end{equation}
For a node inside $\mathcal{D}$ (i.e. $\mathbf{x}_{k,i} \in \mathcal{D}$) a bigger step-size in the update equation is required to move in the direction of $\mathbf{w}^o$. So we replace $\mu_{k,i}$ in \eqref{eqwup} as follows
\begin{equation}
\mu_{k,i}=\alpha \mu_{k,i-1}
\end{equation}
where  $\alpha>1$. Obviously, in this case $\mu_{k,i}>\mu_{k,i-1}$. It should be noted that according to the recursive equation in \eqref{eqwup}, increasing the step size may lead to algorithm divergence. So, in order to avoid algorithm divergence we consider an upper bound for step sizes as
\begin{equation}
\mu_{k,i}=
\begin{cases}
    \alpha \mu_{k,i-1}       & \quad    \alpha \mu_{k,i-1} < \mu_{\max} \\
    \mu_{\max}               & \quad \alpha \mu_{k,i-1} \geq \mu_{\max}    \\
  \end{cases}
  \ \ \ (\mathrm{if}\ \mathbf{x}_{k,i} \in \mathcal{D})
\end{equation}
When $\mathbf{x}_{k,i} \notin \mathcal{D}$ we need to reduce the step size as nodes moves toward the target at $\mathbf{w}^o$. In this case the  step size adaptation function can be given by
\begin{equation}
\mu_{k,i}=\beta \mu_{k,i-1}+\gamma e_{k}^2(i)
\end{equation}
where $e_{k}^2(i)=d_k(i)-\mathbf{u}_{k,i}(\mathbf{w}_{k,i-1}-\mathbf{x}_{k,i-1})$. In this case $\mu_{k,i}$ becomes smaller as node approaches the target. To avoid slow convergence rate, we can consider a lower bound for step size as follows
\begin{equation}
\mu_{k,i}=
\begin{cases}
    \alpha \mu_{k,i-1}       & \quad    \beta \mu_{k,i-1}+\gamma e_{k}^2(i) > \mu_{\min} \\
    \mu_{\min}               & \quad    \beta \mu_{k,i-1}+\gamma e_{k}^2(i) \leq \mu_{\min}    \\
  \end{cases}
  \ \ \ (\mathrm{if}\ \mathbf{x}_{k,i} \notin \mathcal{D})
\end{equation}

As mentioned in the introduction, in some applications, networks cannot afford large communication overhead due to
energy consumption cost or bandwidth restrictions. In order to minimize the communication overhead for this applications,
we must introduce a new metric that considers only a subset of neighbours to consult at node $k$. To this end, we select
neighbors of node $k$ that have small estimated variance product measure and ignore the other neighbors. One way is to change \eqref{ndef}  as
\begin{equation}  \label{nnei}
j \in \mathcal{N}_k\ \ \mathrm{if}\ \ \|\mathbf{x}_{j,i}-\mathbf{x}_{k,i}\|\leq R\ \ \mathrm{and}\ \ \sigma_{j,i}^2 \leq \sigma_{k,i}^2
\end{equation}

\begin{rem}
To use \eqref{nnei}  we need $\sigma_{k,i}^2$  which are unknown in general and for practical usage they must be estimated.  At every node $k$, $\sigma_{k,i}^2$ can be estimated by time-averaging as
\begin{equation}
\hat{\sigma}_{k,i}^2 = \eta \hat{\sigma}_{k,i-1}^2 +(1-\eta) (d_k(i)-\mathbf{u}_{k,i} \mathbf{w}_{k,i-1})
\end{equation}
using a forgetting factor $0<\eta<1$.
\end{rem}
Finally, using the introduced modifications we arrive at the proposed algorithm as given in the Table. 1.

\begin{table*}[t] 
\centering
\caption{Proposed Diffusion Mobile Adaptive Network with Selective Cooperation.}
\def\arraystretch{1.4}
\begin{tabular}[c]{|l|}
\hline
Every node $k$ in the network performs the following steps for $i>0$. \\
The node has access to the local data \\ \hspace{2cm}
$\{{d_k}(i),{\mathbf{u}_{k,i}},{\mathbf{v}_{k,i}},\sigma_{n,k}^2\}$
\\
If $ \|\mathbf{w}_{k,i}-\mathbf{x}_{k,i}\|^2 \gg s$ then  \\ \hspace{2cm}
	$	\mu_{k,i}=
\begin{cases}
    \alpha \mu_{k,i-1}       & \quad    \alpha \mu_{k,i-1} < \mu_{\max} \\
    \mu_{\max}               & \quad \alpha \mu_{k,i-1} \geq \mu_{\max}    \\
  \end{cases}
  \ \ \ (\mathrm{if}\ \mathbf{x}_{k,i} \in \mathcal{D})$  \\
else \\ \hspace{2cm}
$	\mu_{k,i}=
\begin{cases}
    \alpha \mu_{k,i-1}       & \quad    \beta \mu_{k,i-1}+\gamma e_{k}^2(i) > \mu_{\min} \\
    \mu_{\min}               & \quad    \beta \mu_{k,i-1}+\gamma e_{k}^2(i) \leq \mu_{\min}    \\
  \end{cases}
  \ \ \ (\mathrm{if}\ \mathbf{x}_{k,i} \notin \mathcal{D}) $  \\
Use \eqref{nnei} to find set of neighbor nodes of every node $k$ (i.e. $\mathcal{N}_{k,i})$   \\
Compute the following local adaptation and criterion\\ \hspace{2cm}
${{{\bf{m}}}_{k,i}} = {{\bf{w}}_{k,i - 1}} + {\mu _{k}}\sum\limits_{\ell  \in {{\cal N}_k}} c_{\ell ,k}^w{\bf{u}}_{\ell ,i}^T\left( {{d_{\ell ,i}} - {{\bf{u}}_{\ell ,i}}{{\bf{w}}_{k,i - 1}}} \right) $ \\ \hspace{2cm}
${\mathbf{s}}_{k,i} = {\mathbf{v}}_{k,i - 1}^g + {\delta_k}\sum\limits_{\ell  \in {{\cal N}_k}} {c_{\ell ,k}^v\left( {{{\mathbf{v}}_{\ell ,i}} - {\mathbf{v}}_{k,i - 1}^g} \right)}$ \\ 
Perform two local combination steps using data from selected neighbors \\ \hspace{2cm}
${{\bf{w}}_{k,i}} = \sum\limits_{\ell  \in {{\cal N}_k}} a_{\ell ,k}^w{{\bf{m}}_{\ell ,i}}$ \\ \hspace{2cm}
${\mathbf{v}}_{k,i}^g = \sum\limits_{\ell  \in {{\cal N}_k}} {a_{\ell ,k}^v{{{\mathbf{s}}}_{\ell ,i}}}$ \\
Update the node velocity and its location \\ \hspace{2cm}
${\mathbf{v}_{k,i}} = \xi_1 \frac{{\mathbf{w}_{k,i-1}} - {\mathbf{x}_{k,i-1}}}{{\left\| {{\mathbf{w}_{k,i-1}} - {\mathbf{x}_{k,i-1}}} \right\|}} + \xi_2 \mathbf{v}_{k,i-1}^g  + \xi_3 \sum\limits_{l \in {\mathcal{N}_k}\backslash \{k\} }(\|\mathbf{x}_{l,i}-\mathbf{x}_{k,i}\|-r) \frac{\mathbf{x}_{l,i}-\mathbf{x}_{k,i}}{\|\mathbf{x}_{l,i}-\mathbf{x}_{k,i}\|}$   \\ \hspace{2cm}
${\mathbf{x}_{k,i }} = {\mathbf{x}_{k,i-1}} + {\mathbf{v}_{k,i}} \Delta t$ \\
As $i$ evolves we have ${\mathbf{w}_{k,i}} \approx {\mathbf{w}^o},{\mathbf{v}_{k,i} \approx 0}$
\\ \hline
\end{tabular}
\end{table*}

\section{Simulation Results}
 In this section we present simulation results to evaluate the performance of our proposed algorithm. We use a network with $N=50$ nodes that are initially uniformly distributed inside a cube with length 10. Their velocities are set at random directions and unit magnitude.
 The simulation parameters are set as follows. The factors of velocity control are  $\xi_1=0.8$, $\xi_2=0.5$, $\xi_2=0.8$. We further set $\beta=0.85,\ \gamma=0.001$. The combination coefficients are set as  ${a_{l,k}^w}={a_{l,k}^v}={c_{l,k}^w}={c_{l,k}^v}=1/|\mathcal{N}_k(i)|$ if $l \in \mathcal{N}_k(i)$. We set the time duration to $\Delta t=0.5$ and $\delta_=0.5$. Moreover, optimal distance between two neighbors is set to $r=2$. A node chooses nearest neighbors from neighbors within the radius $R=6$. The observation noise $n_k(i)$ is assumed to be zero-mean Gaussian noise which is given by \eqref{noap}. 

\begin{figure}[t]
\centering 
\includegraphics [width=8.5cm]{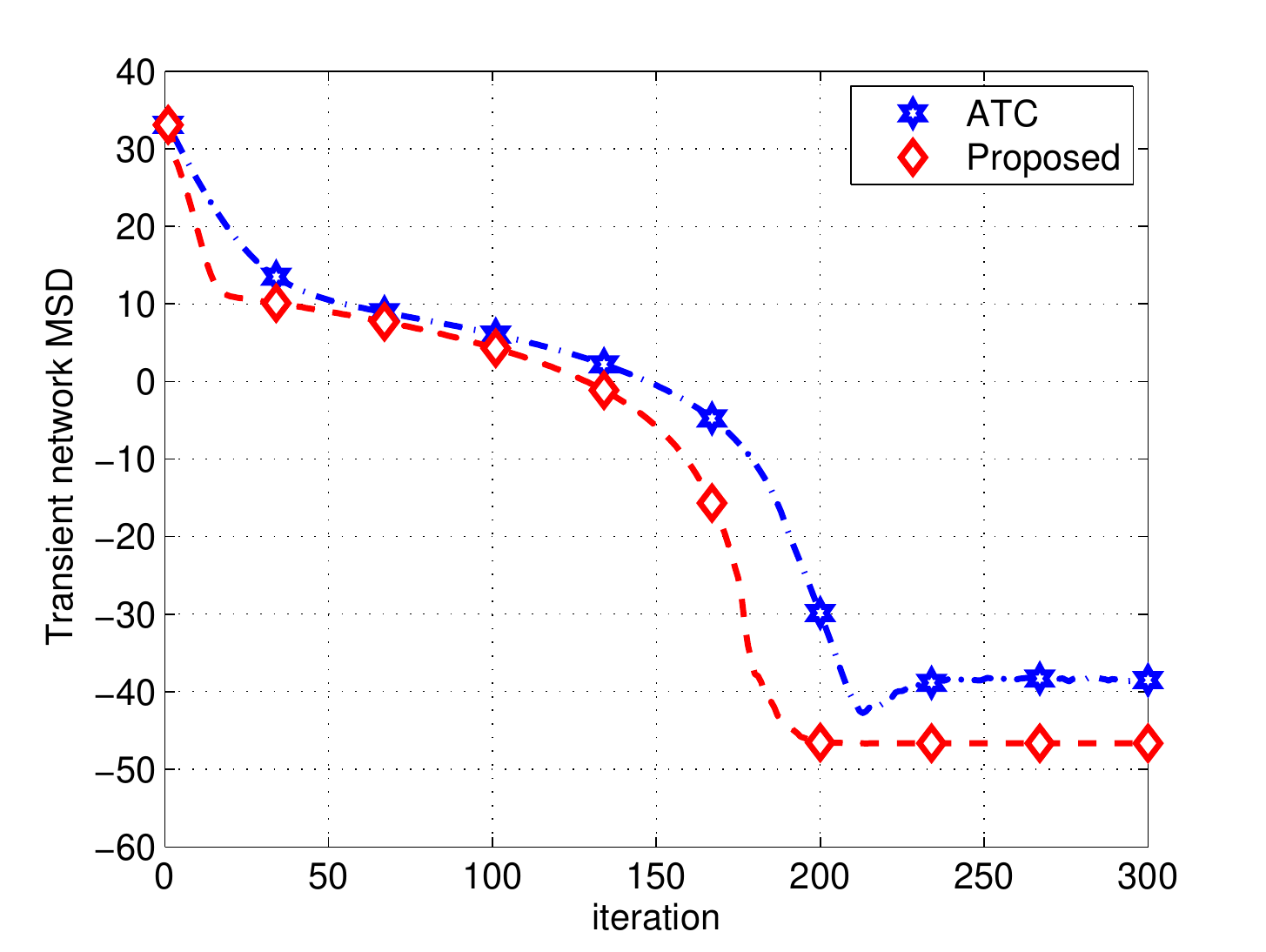} 
\centering\caption{Transient network MSD for estimating the target location at $\mathbf{w}^o=[120\ 120]^T$.}
\label{fig-2}
\end{figure}

Fig. \ref{fig-2} shows the network transient mean-square deviation (MSD) (for two different algorithms) which is defined as
\begin{equation}
\mathrm{MSD}_i=\frac{1}{N} \sum_{k=1}^N E[\|\mathbf{w}^o-\mathbf{w}_{k,i}\|^2]
\end{equation}
In the following, we compare the performance of the proposed algorithm with the ATC algorithm given in \cite{tu11}. Note that for the ATC algorithm  each node cooperates with fixed number of its nearest neighbors at every iteration ($\mathcal{N}_{k,i}=4$). Moreover, each node uses fixed step-size $\mu_k=0.5$ at every iteration. 

\begin{figure}[t]
\centering 
\includegraphics [width=8.5cm]{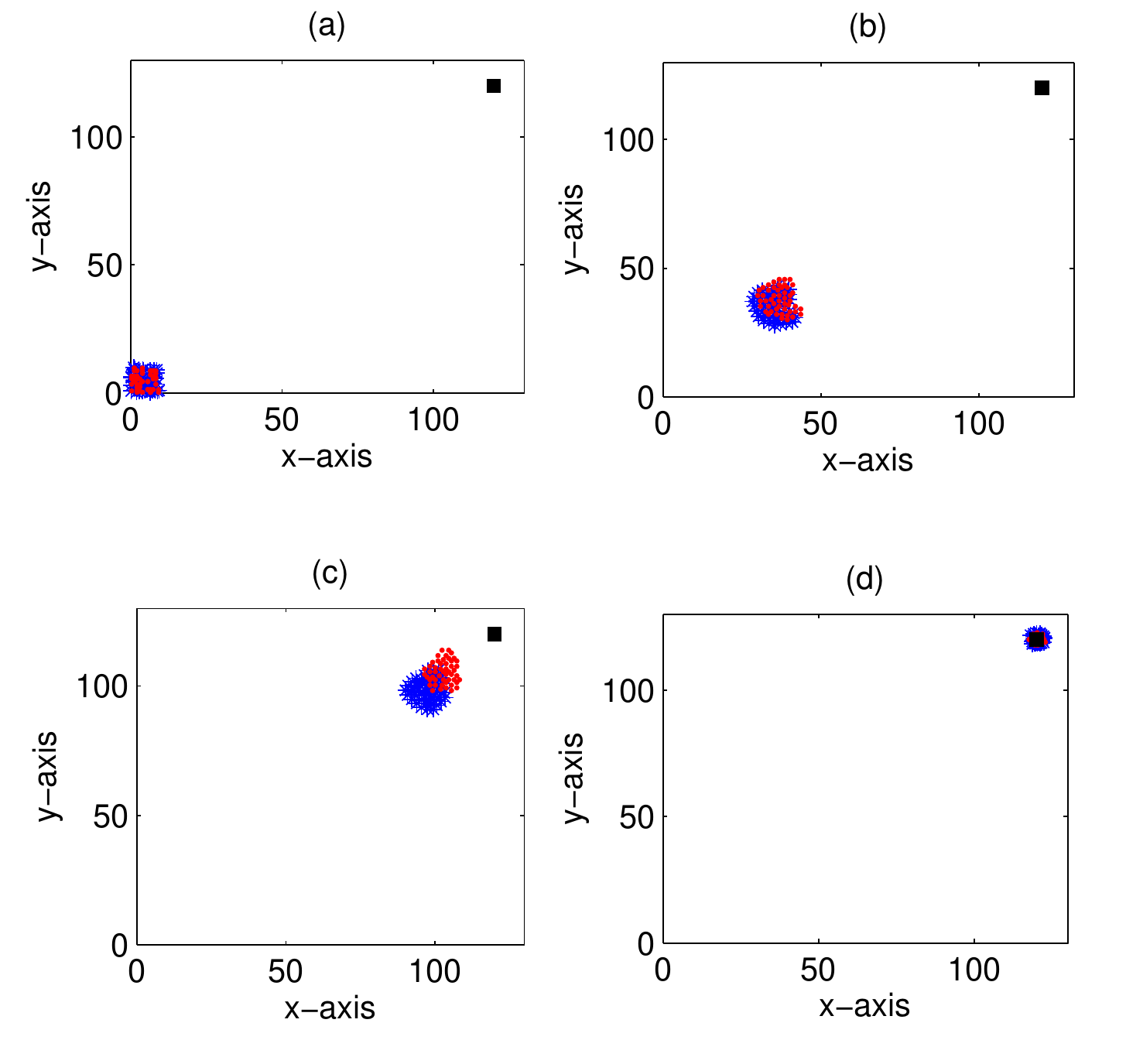} 
\centering\caption{Maneuvers of mobile networks (with the ATC and the proposed algorithms) over time. (a) $i=1$, (b) $i=50$, (c) $i=150$, and (d)  $i=300$. Note that ``*" and  ``$\bullet$" indicates the locations and moving directions of the nodes with ATC, proposed algorithm respectively and ``$\blacksquare$" denotes the location of the target.}
\label{fig-3}
\end{figure}

 Fig. \ref{fig-2} shows the transient network MSD for estimating the target location at $\mathbf{w}^o=[120\ 120]^T$. These curves
are averaged over 50 experiments with the same initial state of $\mathbf{w}_{k,-1}$ for all $k$. We observe
that for both algorithms the transient  MSD decreases dramatically in the first phase, then the network moves towards the target and finally, at steady state, the network arrives at the target. However, the proposed algorithm provides better performance compared to the ATC diffusion algorithm. Fig. \ref{fig-3}  illustrates the maneuver of a mobile network (with the ATC and proposed algorithm) over time. We observe that both algorithms exhibits harmonious movement, but the proposed algorithm moves faster to the target. 

\begin{figure}[t]
\centering 
\includegraphics [width=8.5cm]{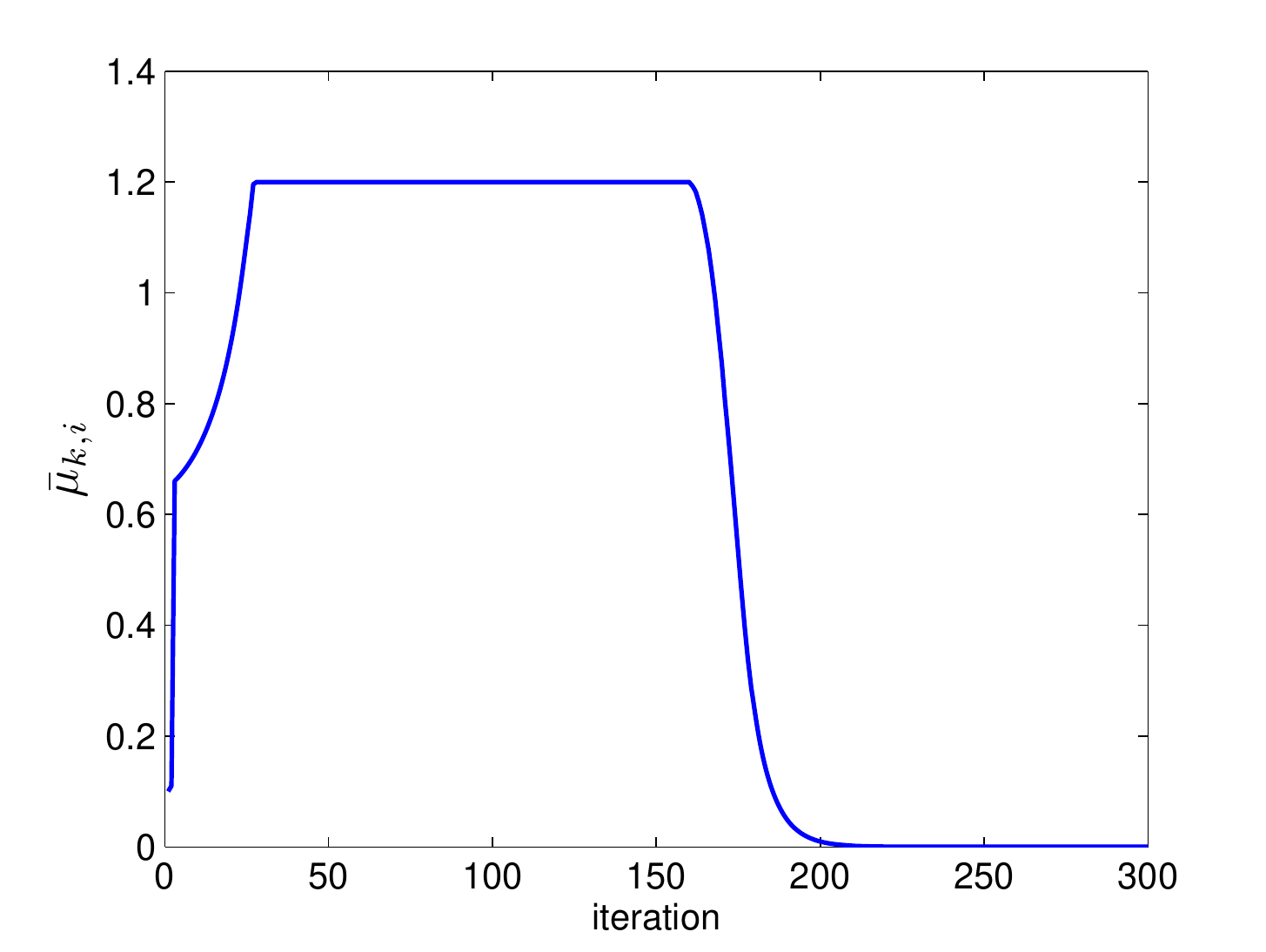} 
\centering\caption{The network average step size $\bar{\mu}_{k,i}$ over iteration.}
\label{fig-4}
\end{figure}

\begin{figure}[t]
\centering 
\includegraphics [width=8.5cm]{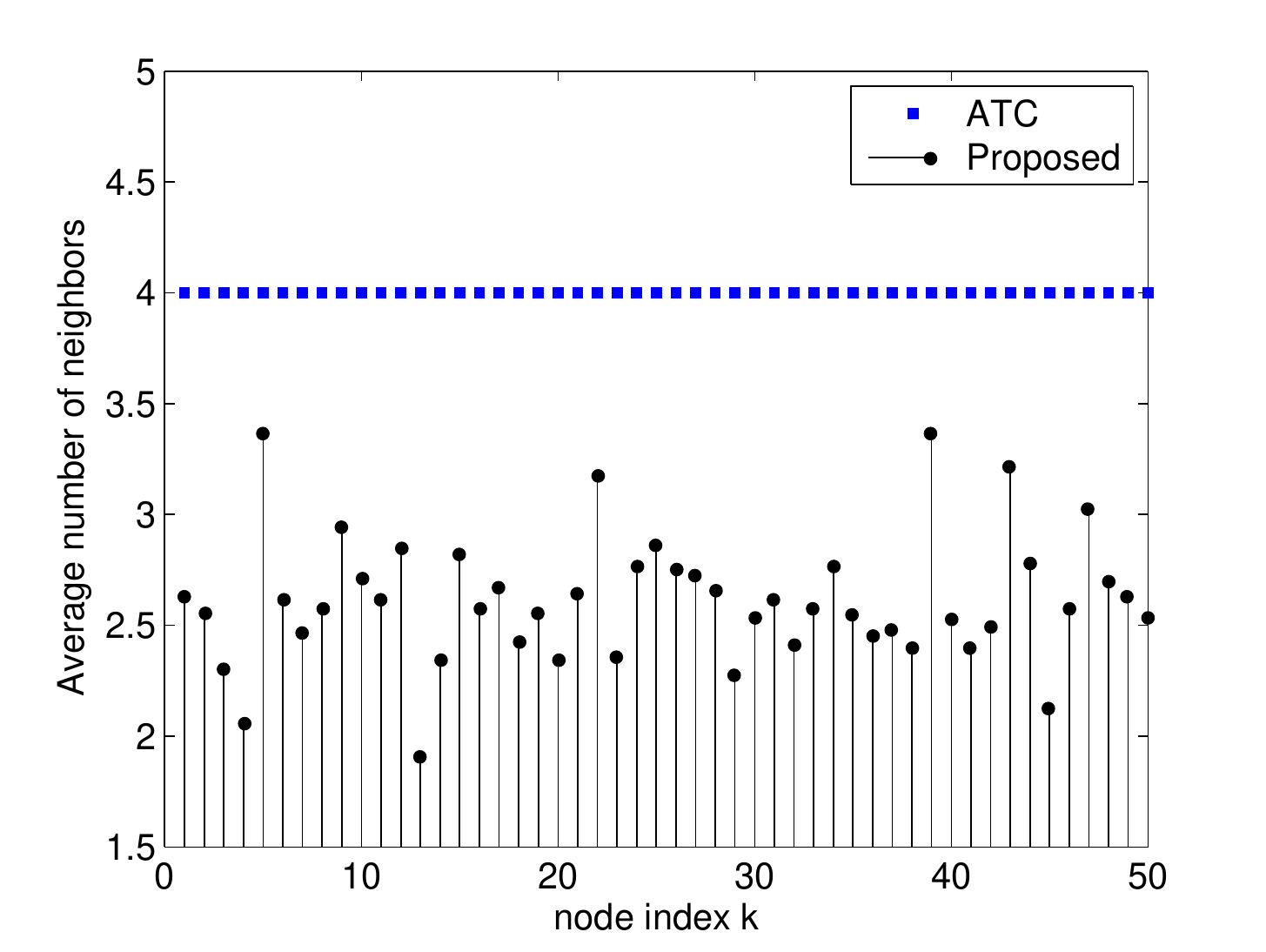} 
\centering\caption{The average number of neighbors for every node.}
\label{fig-5}
\end{figure}

Fig. \ref{fig-4} shows how the network average step size (i.e. $\bar{\mu}_{k,i}=\frac{1}{N}\sum_{k=1}^{N} \mu_{k,i}$ evolves over time. We can see that at the first iteration since the nodes are too far from the target, at every node the step size parameter has been increased iteratively so that after some iterations we have $\mu_{k,i}=\mu_{\max}$. As the nodes move  toward the target, at every node the step size parameter has been decreased iteratively.  
Fig. \ref{fig-5} shows the average number of neighbors for every node. Note that for the ATC algorithm we have $\mathcal{N}_{k,i}=4$ (according to our simulation setup) while the proposed algorithm uses small number of neighbors which means it requires less communication load. 

\section{Conclusions}
In this paper we proposed a modified ATC diffusion algorithm for mobile adaptive networks where the individual agents are
allowed to move in pursuit of a target. The motivation was that in the ATC algorithm fixed step sizes
are used in the update equations for velocity vectors and location vectors. When the nodes are too far away from the target, such
strategies may require large number of iterations to reach the target. To address this issue, in the proposed algorithm we used distance-based variable step
sizes for adjustment at diffusion algorithms to update velocity vectors and location vectors. We also used a selective cooperation where only a subset of nodes at each iteration is used to share information. The performance of the proposed algorithm was evaluated by
simulation tests where the obtained results showed the superior performance of the proposed algorithm in comparison with the available
ATC mobile adaptive network.

\end{document}